\documentclass[runningheads]{llncs}
\usepackage{graphicx}
\usepackage{subfig}
\graphicspath{{imgs/}}
\begin{document}
\title{Symbiotic Hybrid Neural Network Watchdog For Outlier Detection}
\titlerunning{Symbiotic Neural Network Watchdog}
%
\author{Justin Bui\inst{1} \and
Robert J. Marks II\inst{1}}
\authorrunning{J. Bui and R. Marks}
%
\institute{Baylor University, Waco, TX, 76798, USA \\
\email{Justin\_Bui@baylor.edu}\\
\email{Robert\_Marks@baylor.edu}\\
\url{https://www.ecs.baylor.edu/}} 
\maketitle              
\begin{abstract}
 Neural networks are largely black boxes. A neural network trained to classify fruit may classify a picture of a giraffe as a banana. A neural network watchdog's job is to identify such inputs, allowing a classifier to disregard such data. We investigate whether the watchdog should be separate from the neural network or symbiotically attached. We present empirical evidence that the symbiotic watchdog performs better than when the neural networks are disjoint.

\end{abstract}

\section{Introduction}
The neural network watchdog is a tool used to determine whether a classification or regression has been performed on an input that is in-distribution or out-of-distribution with respect to the training data \cite{Bui}. The work focused on the disjoint approach to the watchdog, where the neural network and the watchdog autoencoder are trained separately on the same data. The watchdog network is used to determined the validity of the input data, allowing for the removal of out-of-distribution classification data from the output in parallel to classification. 

To build upon the application of the watchdog, we propose the use of a symbiotic neural network where the autoencoder \cite{Kong,Thompson1,Thompson2} is symbiotically attached to the neural network under scrutiny. This hybrid system is capable of generating and classifying input data without the need for completely separate networks. This allows the watchdog to regenerate input data using identical input weights and bias up to the neural network's inflection point. In our analysis, this allows for more precise watchdog performance since the generation is closely coupled to the initial classification layers. Since the generator and classifier share a number of layers, the symbiotic watchdog exhibits strong performance gains in training, evaluation, and prediction when compared to a disjoint watchdog.

\section{Background}
The precise definition of a {\em hybrid neural network} is open for interpretation and is used in different contexts by different researchers. Work done by McGarry et al. \cite{McGarry} discusses the integration of neural networks into symbolic systems, whereas Yang et al. \cite{Yang} discusses a retrieval-generation models. Hybrid neural networks have been applied in various areas, such as power load forecasts \cite{Abedinia},\cite{Amjady},\cite{Kim}, medical analysis techniques \cite{Dokur}, and financial applications \cite{Lee2,Lee3,Yao}. Their flexibility allows for new and novel techniques to solving modern day problems. The phrase hybrid also extends beyond network structures, to training and evaluation techniques \cite{Kobayashi,Pitiot}.

\section{Proof of Concept}
We  investigate the feasibility of creating a hybrid classifier-generator network to be used with the neural network watchdog where the watchdog shares a portion of the architecture of the neural network. In this sense, the watchdog is symbiotic. Our proof of concept hybrid network is based on a 2D convolutional image classifier and 2D convolutional autoencoder, as described below.

\subsection{Symbiotic Hybrid Network Structure}
\begin{figure}[h!]
	\begin{center}
		\includegraphics[width=.65\textwidth]{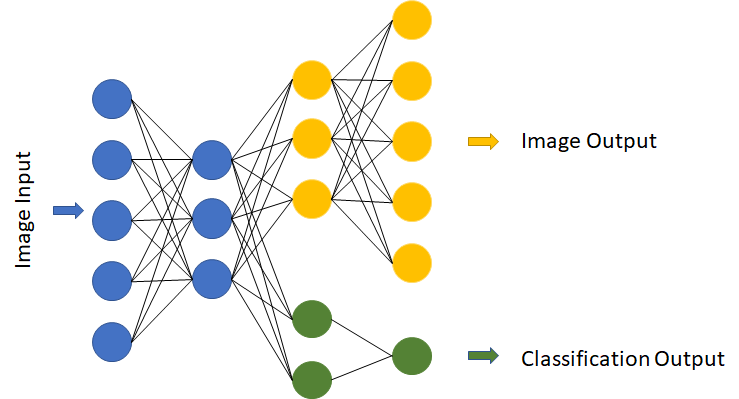}
		\caption{The general structure of a symbiotic hybrid neural network image classifier. This example demonstrates a single image input which splits in to generative and classifying components.}
		\label{HGS}
	\end{center}
\end{figure}

To demonstrate the proof of concept, a symbiotic hybrid convolutional neural network is designed to classify and regenerate MNIST digit images. As seen in Figure~\ref{HGS}, the symbiotic network is comprised of 3 subsections of layers:

\begin{enumerate}
	\item Input Layers
	\item Generative Layers
	\item Classification Layers
\end{enumerate}

\subsubsection{The Input Layers.}
\begin{figure}[h!]
	\begin{center}
		\includegraphics[width=.5\textwidth]{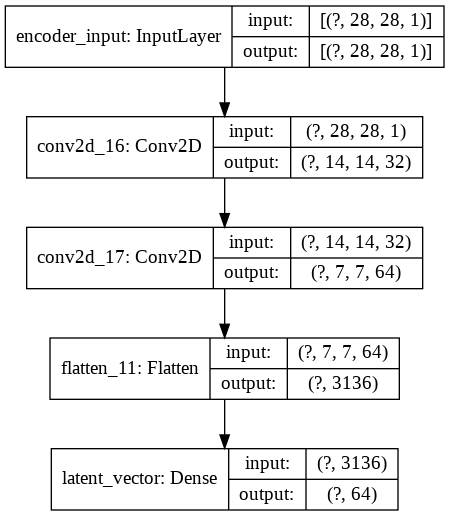}
		\caption{The encoding Layers of the hybrid neural network, which represent the input layers of the symbiotic neural network.}
		\label{ENC}
	\end{center}
\end{figure}

The input layers of the symbiotic hybrid network consist of multiple 2D convolutional layers, as well as a flatten and dense layer, as shown in Figure~\ref{ENC}. These layers comprise the encoding portion of the hybrid network and feeds in to both the generative and classification layers.

\subsubsection{The Generative Layers.}  
\begin{figure}[h!]
	\begin{center}
		\includegraphics[width=.5\textwidth]{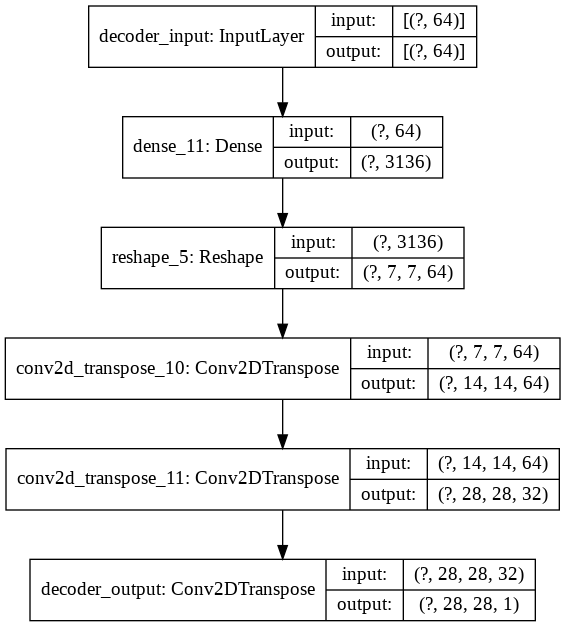}
		\caption{The decoding layers of the symbiotic neural network. When combined with the input layers, as in Figure~\ref{ENC}, create an autoencoder.}
		\label{DEC}
	\end{center}
\end{figure}

As seen in Figure~\ref{DEC}, the generative layers represent the decoder portion of an autoencoder. These layers are responsible for decoding the representation generated by the input layers into an image reconstruction of the input which is used by the Watchdog to determine input validity. 

\subsubsection{The Classification Layers.}
The classification layers convert the input layer encoding into an activation output, producing a distribution corresponding to classification probabilities.

\begin{figure}[h!]
	\begin{center}
		\includegraphics[width=.6\textwidth]{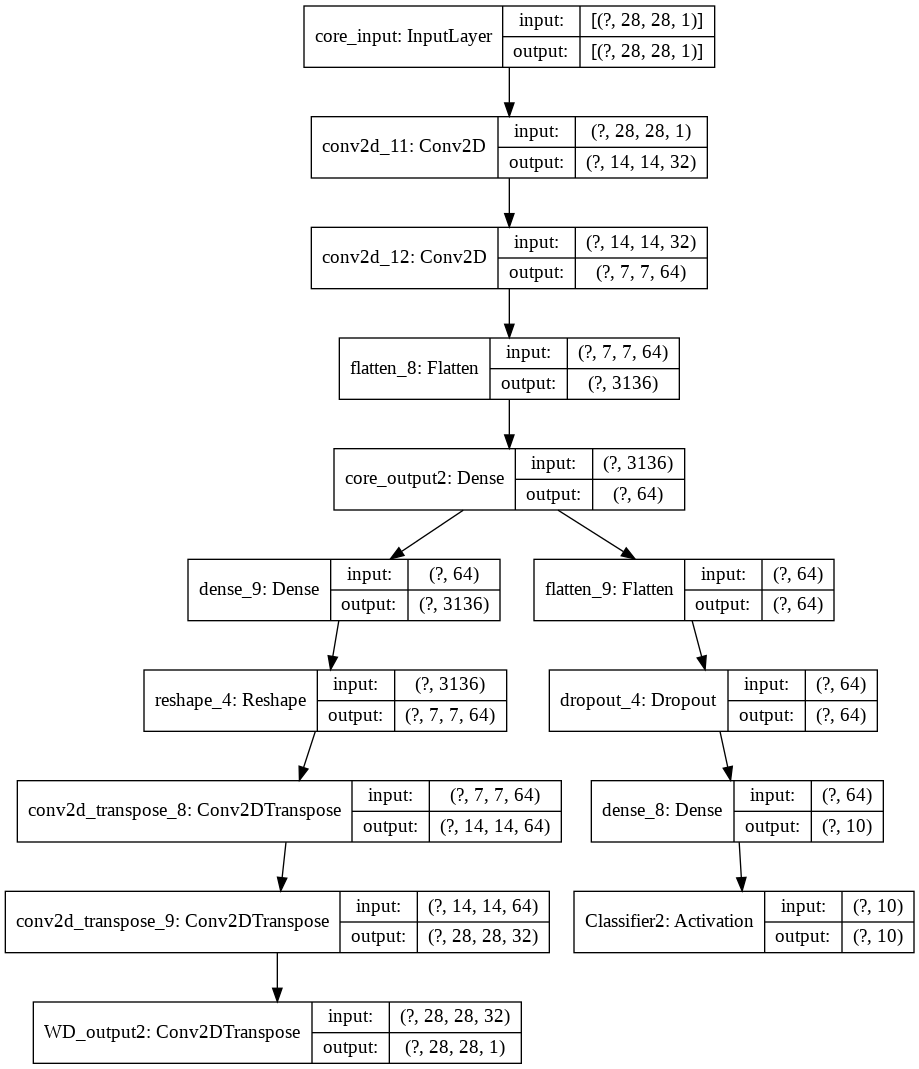}
		\caption{The complete symbiotic neural network structure}
		\label{HYB}
	\end{center}
\end{figure}

The final symbiotic hybrid neural network structure can be seen in Figure~\ref{HYB}.

\subsection{Training The Symbiotic Hybrid Network}

\subsubsection{Training and Evaluation Datasets.}
\begin{figure}[h!]
	\begin{center}
		\includegraphics[width=.5\textwidth]{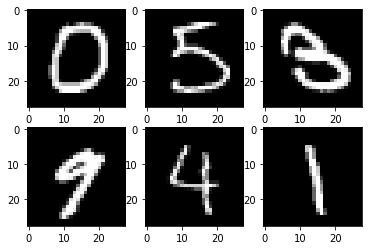}
		\caption{Examples of the in-distribution MNIST Digit dataset.}
		\label{MNI}
	\end{center}
\end{figure}

\begin{figure}[h!]
	\begin{center}
		\includegraphics[width=.5\textwidth]{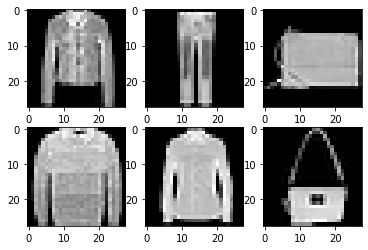}
		\caption{Examples of the out-of-distribution MNIST Fashion dataset.}
		\label{FMN}
	\end{center}
\end{figure}

For this proof of concept, we will be using both the MNIST digit and MNIST Fashion datasets. Our training set is comprised of 60,000 MNIST digit images. The MNIST images are considered in-distribution. In order to evaluate the functionality of the Watchdog, we must also introduce out-of-distribution data, provided by the MNIST Fashion dataset. Examples of the evaluation data may be seen in Figures~\ref{MNI} and \ref{FMN}.

Once the network has been trained, the networks will be evaluated using a mixed-distribution dataset, consisting of 10,000 evaluation images from each of the MNIST digit and MNIST Fashion datasets. 

\subsubsection{Biased Training.}
One of the considerations with training a symbiotic neural network as shown above is the impact of back-propagation bias when dealing with multiple outputs. The network structure introduced in \ref{HGS} will require biased training to improve the performance. Biasing the training weights allows for highly adaptive network performance, depending on the desired outcome of the network. 

To further investigate the importance of the bias, five identical symbiotic neural networks are developed with different bias weights, as well as a sixth independent classifier and autoencoder as a control. The biases for the symbiotic networks are shown in Table~\ref{tab1} below:

\begin{table}\centering
	\caption{Symbiotic hybrid network training weights.}\label{tab1}
	\begin{tabular}{|l|l|l|}
		\hline
		Network & Classifier Weights & Generator Weights \\ 
		\hline
		Classifier Biased & 1.0 & 0.0 \\ 
		\hline
		Generator Biased & 0.0 & 1.0 \\
		\hline
		25\% Class Biased & 0.25 & 0.75 \\ 
		\hline
		50\% Class Biased & 0.5 & 0.5 \\ 
		\hline
		75\% Class Biased & 0.75 & 0.25 \\
		\hline
	\end{tabular}
\end{table}

Evaluating the efficacy of the symbiotic hybrid network can be performed by comparing the RMSE values of the hybrid's classifier and generator results with the RMSE values of the disjointed (independent) watchdog.  

\subsection{Evaluating the Generator}
\begin{figure}[h!]
	\begin{center}
		\includegraphics[width=.5\textwidth]{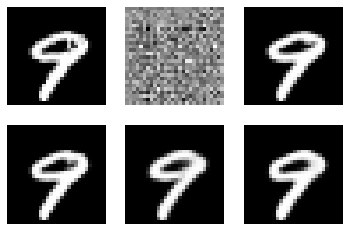}
		\caption{From left to right: Original MNIST image, classifier biased symbiotically generated image, generator biased symbiotically generated image, and independent autoencoder generated image of the MNIST Digit 9.}
		\label{NIN}
	\end{center}
\end{figure}

The generator can be evaluated by calculating the root mean squared error (RMSE) between the original and the generated images. Figure~\ref{NIN} shows examples of an original image, as well as the symbiotic hybrid and independent autoencoder generated images.

\begin{figure}[h!]
	\begin{center}
		\includegraphics[width=.5\textwidth]{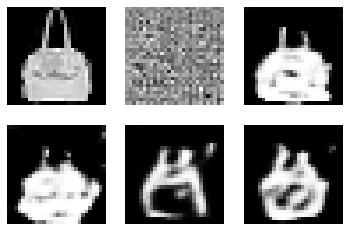}
		\caption{From Left to Right: Original MNIST fashion image, classifier biased symbiotically generated image, generator biased symbiotically generated image, and independent autoencoder generated image of a MNIST Fashion Purse.}
		\label{PUR}
	\end{center}
\end{figure}

\begin{table}[h]\centering
	\caption{RMSE values of each neural network, by image type}\label{tab2}
		\begin{tabular}{ |l|l|l|l|l|l|l| } 
		\hline
		& Independent & Class. Bias  & Gen. Bias & 25\% Class. & 50\% Class. & 75\% Class.  \\ 
		\hline
		MNIST Images &  1.047 & 13.198 & 1.0396 & 1.696 & 1.982 & 2.358 \\ 
		\hline
		Fashion Images & 6.8824 & 11.387 & 6.898 & 7.372 & 6.680 & 8.305 \\
		\hline
	\end{tabular}
\end{table}

The average RMSE values for each of the networks is shown in Table~\ref{tab2}. These values indicate minor variations in the performance of the generative components of the symbiotic and the stand-alone autoencoders. These variations are expected, as the training weights of the symbiotic networks are adjusted by both the classifier and the generator outputs.

\subsection{Evaluating the Classifier}
The classifier is measured by the evaluation dataset accuracy, as well as examining the ROC curves. Results for the classification accuracy Evaluation of the networks on the MNIST dataset can be seen in Table~\ref{tab3}.

\begin{table}[h]\centering
	\caption{MNIST Classifier Accuracy}\label{tab3}
	\begin{tabular}{ |l|l| } 
		\hline
		& Accuracy \\ 
		\hline
		CNN &  98.81\% \\
		\hline
		Classifier Bias  & 98.83\% \\
		\hline
		Generator Bias & 11.92\% \\
		\hline
		25\% Class Bias & 97.88\% \\
		\hline
		50\% Class Bias & 98.43\% \\
		\hline
		75\% Class Bias & 98.52\% \\
		
		\hline
	\end{tabular}
\end{table}

\begin{figure}[h!]
	\begin{center}
		\includegraphics[width=.65\textwidth]{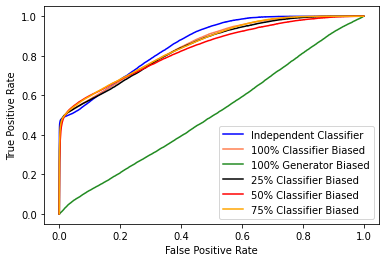}
		\caption{Normalized ROC plots for the unguarded performance of all six classification networks.}
		\label{ROC}
	\end{center}
\end{figure}
\subsection{Symbiotic vs. Independent Networks}

\begin{figure}[h!]
	\begin{center}
		\includegraphics[width=.65\textwidth]{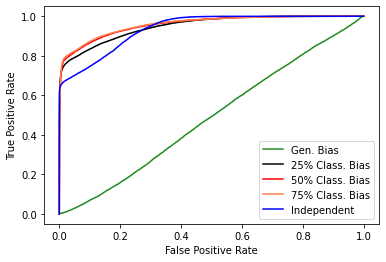}
		\caption{Normalized ROC curves for the watchdog guarded symbiotic and independent classifiers. The RMSE threshold for these curves is set to 6.5.}
		\label{ROC2}
	\end{center}
\end{figure}

\subsubsection{Training and Evaluation Execution Times.}  
In addition to evaluating the classification performance of both the independent and symbiotic networks, a comparison of training and evaluation times is performed. For this evaluation, the following training parameters are used: 60,000 MNIST digit images for training, with 10 training epochs. These measurements are performed on both GPU and CPU runtime environments using Google's Colab notebooks. Table~\ref{tab4} displays the training time breakdown for the networks.  

\begin{table}[h]\centering
	\caption{Training times for the networks.}\label{tab4}
	\begin{tabular}{ |l|l|l| } 
		\hline
		& GPU  Runtime & CPU Runtime  \\ 
		\hline
		Independent Watchdog: & 638.3s & 2408.5s  \\ 
		\hline
		Classifier Biased: & 553.8s & 1692.4s \\
		\hline
		Generator Biased: & 567.5s & 1688.9s \\
		\hline
		25\% Classifier Bias: & 552.8s & 1706.8s \\
		\hline
		50\% Classifier Bias: & 564.3s & 1727.6s \\
		\hline
		75\% Classifier Bias: & 567.3s & 1747.0s \\
		\hline
	\end{tabular}
\end{table}

Similar to the improved performance with regards to training time, the evaluation times for the networks has been measured, and improved performance is found with the symbiotic watchdog. For the evaluation parameters, all six networks are evaluated using the 20,000 digit mixed-distribution dataset.

\begin{table}[h]\centering
	\caption{Evaluation times for the networks.}\label{tab5}
	\begin{tabular}{ |l|l|l| } 
		\hline
		& GPU  Runtime & CPU Runtime  \\ 
		\hline
		Independent Watchdog: & 1.51s & 20.89s  \\ 
		\hline
		Classifier Biased: & 0.984s & 14.81s \\
		\hline
		Generator Biased: & 0.964s & 14.99s \\
		\hline
		25\% Classifier Bias: & 0.951s & 14.97s \\
		\hline
		50\% Classifier Bias: & 0.968s & 14.97s \\
		\hline
		75\% Classifier Bias: & 0.933s & 15.13s \\
		\hline
	\end{tabular}
\end{table}

\section{Conclusion}

A symbiotic autoencoder watchdog is developed in conjunction with a symbiotic  generator/classification neural network. Creating a hybrid classification neural network, as demonstrated here, results in better training and evaluation performance, as seen in Tables \ref{tab3}, \ref{tab4}, and \ref{tab5}. Classification performance closely matches the performance of an independent watchdog, as can be seen in Figures \ref{ROC} and \ref{ROC2}. The choice of an RMSE threshold is ultimately determined by the specific application. Results may vary based on data and the trade off between detection and false alarms.


\begin{thebibliography}{00}
	
	\bibitem{Abedinia} Abedinia, Oveis, Nima Amjady, and Noradin Ghadimi. ``Solar energy forecasting based on hybrid neural network and improved metaheuristic algorithm." Computational Intelligence 34.1 (2018): 241-260.
	
	\bibitem{Amjady} Amjady, N., and S. A. Banihashemi. ``Transient stability prediction of power systems by a new synchronism status index and hybrid classifier." IET generation, transmission \& distribution 4.4 (2010): 509-518.
	
	\bibitem{An} An, Ning, et al. ``Using multi-output feedforward neural network with empirical mode decomposition based signal filtering for electricity demand forecasting." Energy 49 (2013): 279-288.
	
	\bibitem{Beck} Beck, Steven, et al. ``A hybrid neural network classifier of short duration acoustic signals." IJCNN-91-Seattle International Joint Conference on Neural Networks. Vol. 1. IEEE, 1991.
	
	\bibitem{Bui} Bui, Justin, and Robert J. Marks II. ``Autoencoder Watchdog Outlier Detection for Classifiers." arXiv preprint arXiv:2010.12754 (2020).
	
	\bibitem{Curteanu} Curteanu, Silvia, and Florin Leon. ``Hybrid neural network models applied to a free radical polymerization process." Polymer-Plastics Technology and Engineering 45.9 (2006): 1013-1023.
	
	\bibitem{Dokur} Dokur, Zümray, and Tamer Ölmez. ``ECG beat classification by a novel hybrid neural network." Computer methods and programs in biomedicine 66.2-3 (2001): 167-181.
	
	\bibitem{Hernandez} Hernández, Gerardo, et al. ``Hybrid neural networks for big data classification." Neurocomputing 390 (2020): 327-340.
	
	\bibitem{Jirapummin} Jirapummin, Chaivat, Naruemon Wattanapongsakorn, and Prasert Kanthamanon. ``Hybrid neural networks for intrusion detection system." Proc. of ITC–CSCC. Vol. 7. 2002.
	
	\bibitem{Kim} Kim, Myoungsoo, et al. ``A hybrid neural network model for power demand forecasting." Energies 12.5 (2019): 931.
	
	\bibitem{Kobayashi} Kobayashi, Takahisa, and Donald L. Simon. ``Hybrid neural-network genetic-algorithm technique for aircraft engine performance diagnostics." Journal of Propulsion and Power 21.4 (2005): 751-758.
	
	\bibitem{Kong} Kong, Gyuyeol, Minchae Jung, and Visa Koivunen. ``Waveform Recognition in Multipath Fading using Autoencoder and CNN with Fourier Synchrosqueezing Transform."  2020 IEEE International Radar Conference (RADAR), pp. 612-617. IEEE, 2020.
	
	\bibitem{Lee} Lee, Eric Wai Ming, et al. ``A hybrid neural network model for noisy data regression." IEEE Transactions on Systems, Man, and Cybernetics, Part B (Cybernetics) 34.2 (2004): 951-960.
	
	\bibitem{Lee2} Lee, Kun Chang, Ingoo Han, and Youngsig Kwon. ``Hybrid neural network models for bankruptcy predictions." Decision Support Systems 18.1 (1996): 63-72.
	
	\bibitem{Lee3} Lee, Tian-Shyug, et al. ``Credit scoring using the hybrid neural discriminant technique." Expert Systems with applications 23.3 (2002): 245-254.
	
	\bibitem{Lim} Lim, Chee-Peng, Jenn-Hwai Leong, and Mei-Ming Kuan. ``A hybrid neural network system for pattern classification tasks with missing features." IEEE transactions on pattern analysis and machine Intelligence 27.4 (2005): 648-653.
	
	\bibitem{McGarry} McGarry, Kenneth, Stefan Wermter, and John MacIntyre. ``Hybrid neural systems: from simple coupling to fully integrated neural networks.'' Neural Computing Surveys 2.1 (1999): 62-93.
	
	\bibitem{Pitiot} Pitiot, Alain, et al. ``Texture based MRI segmentation with a two-stage hybrid neural classifier.'' Proceedings of the 2002 International Joint Conference on Neural Networks. IJCNN'02 (Cat. No. 02CH37290). Vol. 3. IEEE, 2002.
	
	\bibitem{Sun} Sun, Xuli, et al. ``Hybrid neural conditional random fields for multi-view sequence labeling.'' Knowledge-Based Systems 189 (2020): 105151.
	
	\bibitem{Thompson1} Thompson, Benjamin B., Robert J Marks II , Jai J Choi, Mohamed A El-Sharkawi ``Implicit Learning in Autoencoder Novelty Assessment,'' Proceedings of the 2002 International Joint Conference on Neural Networks, 2002 IEEE World Congress on Computational Intelligence, May 12-17, 2002, Honolulu, pp. 2878-2883.
	
	\bibitem{Thompson2} Thompson, Benjamin B., Robert J. Marks II, and Mohamed A. El-Sharkawi ``On the Contractive Nature of Autoencoders: Application to Missing Sensor Restoration,'' 2003 International Joint Conference on Neural Networks, July 20-24, 2003, Portland, Oregon (pp. 3011-3016)
	
	\bibitem{Tsai} Tsai, Chih-Fong, and Yu-Hsin Lu. ``Customer churn prediction by hybrid neural networks.'' Expert Systems with Applications 36.10 (2009): 12547-12553.
	
	\bibitem{Yang} Yang, Liu, et al. ``A hybrid retrieval-generation neural conversation model.'' Proceedings of the 28th ACM International Conference on Information and Knowledge Management. 2019.
	
	\bibitem{Yao} Yao, Ping. ``Hybrid classifier using neighborhood rough set and SVM for credit scoring.'' 2009 International conference on business intelligence and financial engineering. IEEE, 2009.
	
	\bibitem{Zhang} Zhang, Kai, et al. ``Data augmentation for motor imagery signal classification based on a hybrid neural network.'' Sensors 20.16 (2020): 4485.
	
\end{thebibliography}
\end{document}